\pdfoutput=1
\documentclass[11pt]{article}

\usepackage[letterpaper,margin=0.82in]{geometry}
\usepackage[T1]{fontenc}
\usepackage{lmodern}
\usepackage[scaled=0.94]{helvet}
\usepackage{microtype}
\usepackage{amsmath,amssymb,mathtools}
\usepackage{booktabs,multirow,tabularx,array}
\usepackage{graphicx}
\usepackage{xcolor}
\usepackage{enumitem}
\usepackage{xspace}
\usepackage{tikz}
\usetikzlibrary{arrows.meta,positioning,fit,calc}
\usepackage[most]{tcolorbox}
\usepackage{caption}
\usepackage{subcaption}
\usepackage{placeins}
\usepackage{float}
\usepackage{hyperref}
\usepackage[nameinlink,capitalise]{cleveref}

\definecolor{mblue}{HTML}{2364AA}
\definecolor{mteal}{HTML}{2A9D8F}
\definecolor{morange}{HTML}{F4A261}
\definecolor{mred}{HTML}{D95D5D}
\definecolor{mgray}{HTML}{F4F6F8}
\hypersetup{
  colorlinks=true,
  linkcolor=mblue,
  citecolor=mblue,
  urlcolor=mblue,
  pdftitle={MOSAIC: Query-Aware Exploration Policy Adaptation for GraphRAG},
  pdfauthor={EunKyeong Lee, Kyeong-Jin Oh, Jinwon Kim, Hye Woo Lee, Minsang Song, Hyeongjun Jang, Junyoung Youn},
  pdfsubject={Technical Report},
  pdfkeywords={GraphRAG, retrieval-augmented generation, adaptive retrieval, graph retrieval}
}
\setlist[itemize]{leftmargin=1.3em,itemsep=2pt,topsep=4pt}
\setlist[enumerate]{leftmargin=1.5em,itemsep=2pt,topsep=4pt}

\newcommand{\method}{\textsc{Mosaic}\xspace}
\newcommand{\code}[1]{\texttt{#1}}

\title{%
  {\sffamily\bfseries\fontsize{14}{17}\selectfont MOSAIC: Query-Aware Exploration Policy Adaptation for GraphRAG}\\[7pt]
  {\sffamily\large Technical Report}%
}
\author{\sffamily EunKyeong Lee, Kyeong-Jin Oh, Jinwon Kim, Hye Woo Lee\\
\sffamily Minsang Song, Hyeongjun Jang, Junyoung Youn\\[2pt]
\sffamily\small KT Corporation}
\date{\sffamily September 2026}

\begin{document}
\maketitle

\begin{abstract}
Graph Retrieval-Augmented Generation (GraphRAG) can connect evidence distributed across a corpus graph, but most systems execute a largely shared exploration procedure for every query. This creates a structural mismatch: a direct fact may require a compact local neighborhood, a comparison requires balanced coverage of multiple targets, and a mediated question may require a deeper path through a weakly query-related connector. We present \method, a training-free framework that formulates GraphRAG retrieval as a per-query control problem. An LLM analyzer translates the evidence requirements implied by a query into a bounded, executable policy over seed selection, graph traversal, cumulative-gain stopping, and evidence selection. The corpus graph, indexes, scoring functions, grounding procedure, and answer generator remain shared. The current analyzer uses six representative considerations---Mediatedness, Output Cardinality, Seed Coverage, Anchoring Need, Identity--Relation Dependence, and Completeness---as composable reasoning signals rather than mutually exclusive query classes or a closed taxonomy.

On GraphRAG-Bench, \method achieves query-weighted Answer Correctness of 76.97 on Medical and 64.33 on Novel, improving over the strongest previously reported overall results by 5.13 and 4.43 points, respectively. On Medical, it reaches 95.1 Evidence Recall and 86.1 Context Relevancy. Controlled comparisons on an identical graph and generator show that no fixed narrow, medium, or wide policy is consistently optimal; \method improves by 9.96 points over the strongest canonical fixed policy. Relative to Fixed Wide, it evaluates 81.9\% fewer paths and retains 47.2\% fewer evidence items, although its separate analyzer call increases end-to-end latency. Transfer experiments on HotpotQA, MuSiQue, and 2WikiMultiHopQA further show that the policy interface can be applied without benchmark-specific retriever training. These results support analyzer-driven, query-specific exploration as a general and extensible design principle for GraphRAG.
\end{abstract}

\begin{tcolorbox}[colback=mgray,colframe=mblue,title=Technical-report scope]
This report consolidates the method, implementation details, controlled analyses, transfer experiments, and qualitative cases in one document. \method\ is \emph{not} a six-rule retrieval system: it is an analyzer-driven framework for instantiating query-specific retrieval policies. The six considerations in the current implementation are representative and extensible design signals derived from observed retrieval failures.
\end{tcolorbox}

\section{Introduction}

Retrieval-Augmented Generation (RAG) grounds language-model outputs in external evidence \cite{lewis2020rag}. GraphRAG extends this idea by representing entities, relations, passages, or communities as a graph and retrieving connected evidence for questions that require relation following, multi-hop reasoning, or synthesis across sources \cite{edge2024graphrag,guo2024lightrag}. The graph provides useful structure, but it also creates a control problem: retrieval must decide where to start, how far and broadly to traverse, when to stop, and which paths to preserve.

These decisions are usually configured globally. The query changes similarity scores or initial nodes, yet the main operating limits---seed count, depth, width, stopping behavior, and final evidence budget---remain largely shared. A global policy is attractive because it is simple, but it assumes that heterogeneous questions require comparable evidence structures. That assumption is often false. A narrowly stated fact can be diluted by broad exploration; a comparison can fail if seeds cover only one target; and a mediated question can be unreachable under a shallow beam even when the answer-bearing region exists in the graph.

\method\ replaces the search for one globally optimal configuration with explicit per-query policy construction. Before retrieval, an analyzer interprets the question and emits bounded controls for the shared pipeline. During retrieval, cumulative structural gain determines whether the current evidence has saturated, and role-aware selection can preserve complementary paths. Answer generation is held fixed: the analyzer changes what evidence is retrieved, not how the final answer is freely generated.

The contribution is therefore not the observation that queries influence retrieval; many prior systems are query-conditioned, and recent systems adapt routes, edges, constraints, or operators. The technical distinction is the representation and target of adaptation:

\[
\text{query}\;\longrightarrow\;\text{explicit multi-stage retrieval policy}\;\longrightarrow\;\text{shared corpus-graph pipeline}.
\]

The policy jointly controls seed breadth and formulation, traversal depth and width, anchor and connector behavior, stopping sensitivity, and evidence retention. This interface is training-free, validated before execution, and extensible to new evidence requirements.

Our main findings are:

\begin{itemize}
  \item No fixed exploration scope is uniformly effective. On Medical, Fixed Medium reaches 67.01\% ACC, while both narrower (65.45\%) and wider (66.35\%) exploration are worse.
  \item Query-specific control substantially improves retrieval and generation. \method\ obtains 76.97\% overall ACC on Medical and 64.33\% on Novel, with Evidence Recall of 95.1\% and 90.2\%.
  \item The gain is not explained by exhaustive search. Relative to Fixed Wide, \method\ evaluates 81.9\% fewer paths and retains 47.2\% fewer evidence items.
  \item The analyzer emits diverse and requirement-aligned policies. For example, connector traversal is activated for 65.6\% of multi-hop queries but 21.6\% of other queries.
  \item The framework transfers without benchmark-specific retriever training to three standard multi-hop QA benchmarks, while leaving room for stronger answer-format calibration.
\end{itemize}

\section{Related Work}

RAG combines parametric generation with non-parametric retrieval \cite{lewis2020rag,karpukhin2020dpr}. GraphRAG systems retrieve entities, relations, paths, passages, or graph communities: Microsoft GraphRAG supports local and global community-oriented search \cite{edge2024graphrag}; LightRAG combines entity- and relation-level retrieval \cite{guo2024lightrag}; HippoRAG propagates query relevance through personalized PageRank \cite{gutierrez2024hipporag}; and PathRAG prunes relational paths using flow-based signals \cite{chen2026pathrag}. Learned graph retrievers such as GFM-RAG and G-Reasoner encode textual and structural relevance with pretrained graph models \cite{luo2025gfmrag,luo2025greasoner}. Adaptive textual RAG methods decide whether or when to retrieve \cite{jiang2023flare,asai2024selfrag,jeong2024adaptiverag}. Closer graph systems adapt the retrieval paradigm, query-side evidence graph, path constraints, or operator composition. EA-GraphRAG routes between dense and graph retrieval \cite{dong2026eagraphrag}; Relink constructs a query-driven evidence graph on the fly \cite{huang2026relink}; DOTRAG generates query-conditioned constraints for path exploration \cite{moore2026dotrag}; and PAGE-RAG composes heterogeneous evidence operators under bounded budgets \cite{chen2026pagerag}. \method\ does not claim that query-aware GraphRAG is itself new. It differs by constructing an explicit stage-wise operating policy inside one shared corpus-graph retriever, jointly controlling seeds, traversal, stopping, and retained evidence without training an additional router or graph retriever.

\begin{table}[t]
\centering
\caption{High-level location of query adaptation in representative systems. ``Shared'' means globally configured at inference time.}
\label{tab:related}
\small
\begin{tabularx}{\textwidth}{@{}lXXXc@{}}
\toprule
Method & Query-dependent element & Main adaptation target & Shared element & Extra training \\
\midrule
GraphRAG / LightRAG & Entity, relation, or community relevance & Initial retrieval mode or candidates & Expansion limits and budgets & No \\
HippoRAG & PPR reset distribution & Graph personalization & Propagation rule and top-$k$ & No \\
G-Reasoner & Learned query--graph representation & Node/evidence scoring & Learned inference architecture & Yes \\
DOTRAG & Entity-type constraints and path acceptance & Admissible subgraph and iterative paths & Hop/iteration limits & No \\
PAGE-RAG & Query profile & Operator composition and evidence-type budgets & Operator implementations & No \\
\textbf{\method} & Evidence requirements & \textbf{Joint seed, traversal, stop, and evidence policy} & \textbf{Graph, indexes, pipeline, generator} & \textbf{No} \\
\bottomrule
\end{tabularx}
\end{table}

\section{Problem Formulation}

Let $G=(V,E)$ be a corpus graph constructed once from a document collection, and let $q$ be a user query. A conventional graph retriever applies a globally selected configuration $\bar\pi$:
\[
\hat y_q = \mathcal{M}_{\mathrm{ans}}\!\left(q,\mathcal{R}(G,q;\bar\pi)\right).
\]
The objective of \method\ is to replace $\bar\pi$ with a query-specific but bounded policy $\pi_q$ while keeping $G$, the retrieval implementation $\mathcal{R}$, and the answer model $\mathcal{M}_{\mathrm{ans}}$ shared:
\begin{equation}
\pi_q=\mathcal{A}(q)=\left(\pi_q^{\mathrm{seed}},\pi_q^{\mathrm{trav}},\pi_q^{\mathrm{filter}}\right).
\label{eq:policy}
\end{equation}
The policy is generated only from the question. Gold answers, gold evidence, benchmark category labels, and evaluation annotations are not available to the analyzer. All categorical outputs are checked against an allowlist, numerical values are clamped to valid ranges, and invalid or missing fields fall back to conservative defaults.

This formulation separates three concepts that are easily conflated:

\begin{enumerate}
  \item \textbf{Query conditioning}: the query changes similarity or relevance scores.
  \item \textbf{Query classification}: the query is assigned to one of several predefined types, each linked to a fixed pipeline.
  \item \textbf{Query-specific policy construction}: the query is translated into composable stage-level controls executed by a common pipeline.
\end{enumerate}

\method\ implements the third. Diagnostic labels may be logged for analysis or defensive fallback, but they do not select separate retrievers.

\section{From Retrieval Failures to Policy Signals}

We examined retrieval traces in which relevant evidence was present or reachable but the shared policy produced an incorrect answer. The recurring failures did not define six mutually exclusive question types. Instead, they exposed six questions the analyzer should ask when configuring retrieval. \Cref{tab:signals} summarizes the current instantiation.

\begin{table}[t]
\centering
\caption{Representative evidence-requirement signals and their operational effects. Multiple rows may apply to one query.}
\label{tab:signals}
\small
\begin{tabularx}{\textwidth}{@{}p{0.17\textwidth}XXp{0.22\textwidth}@{}}
\toprule
Signal & Analyzer question & Typical fixed-policy failure & Main controls \\
\midrule
Mediatedness & Is the answer reachable only through an unstated intermediate concept? & A weakly related mediator is pruned before the answer-bearing region. & Depth, connector, anchor, stop \\
Output Cardinality & Is one precise value sufficient, or must several items be covered? & Filtering preserves only the highest-scoring answer item. & Seed/evidence budgets, floor, reservation \\
Seed Coverage & Must several targets or facets be initialized separately? & Full-query seeding concentrates on the most salient target. & Per-target query mode, seed budget, balancing \\
Anchoring Need & Must exploration remain tied to an exact entity or subtype? & Traversal drifts to a neighboring but semantically similar concept. & Anchor strength, hub suppression, path mode \\
Identity--Relation Dependence & Is exact entity identity or relation semantics more decisive? & A plausible relation path contains the wrong entity. & Relation-only vs. hybrid scoring \\
Completeness & Is current evidence sufficient, partially missing, or globally misaligned? & A fixed depth stops too early or continues after useful evidence saturates. & Low-gain stop, evidence floor, expansion \\
\bottomrule
\end{tabularx}
\end{table}

The six signals are deliberately not a closed taxonomy. Another corpus, graph schema, domain, or pipeline may expose a new failure pattern. The framework can incorporate it by adding a reasoning cue and mapping it to an existing or newly introduced bounded control. The core contribution is the analyzer-to-policy interface, not the number or names of the present cues.

\subsection{Why controls must be composed}

Individual controls are not independent treatments. A comparison may simultaneously require per-target seeds, connector traversal, and a larger evidence budget. A mediated question may require greater depth but weaker anchoring so that an apparently generic connector is not pruned. Strong anchoring can reduce drift, yet the same anchor can block a necessary mediator. Similarly, stopping quality depends on the evidence produced by the seed and traversal stages. Therefore, control-specific counterfactuals are diagnostic rather than an additive decomposition of total performance.

\section{MOSAIC}
\FloatBarrier

\subsection{End-to-end architecture}

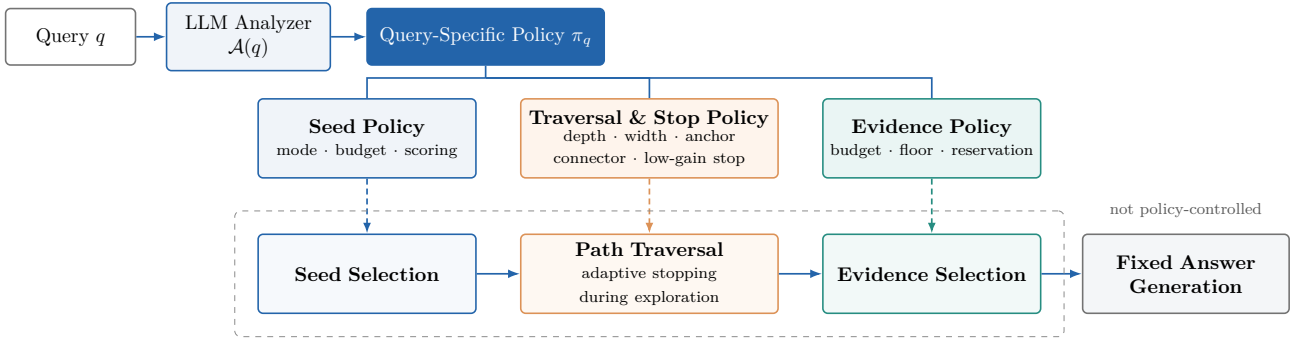
\begin{figure}[H]
\centering
\resizebox{0.98\textwidth}{!}{%
\begin{tikzpicture}[
  box/.style={draw,rounded corners=2.5pt,align=center,font=\small,fill=white,thick},
  input/.style={box,minimum width=23mm,minimum height=10mm,draw=black!55},
  analyzer/.style={box,minimum width=29mm,minimum height=12mm,draw=mblue,fill=mblue!7},
  policy/.style={box,minimum height=14mm,text width=36mm},
  stage/.style={box,minimum height=14mm,text width=36mm},
  arr/.style={-{Latex[length=2.1mm]},line width=0.85pt,mblue},
  control/.style={-{Latex[length=1.9mm]},line width=0.85pt,densely dashed},
  label/.style={font=\scriptsize,text=black!65,align=center}
]
\node[input] (q) at (0,2.15) {Query $q$};
\node[analyzer] (a) at (3.15,2.15) {LLM Analyzer\\$\mathcal{A}(q)$};
\node[box,draw=mblue,fill=mblue,text=white,minimum width=42mm,minimum height=10mm]
  (pi) at (7.35,2.15) {Query-Specific Policy $\pi_q$};
\draw[arr] (q) -- (a);
\draw[arr] (a) -- (pi);

\node[policy,draw=mblue,fill=mblue!7] (ps) at (5.25,0.35)
  {\textbf{Seed Policy}\\[-1pt]\scriptsize mode $\cdot$ budget $\cdot$ scoring};
\node[policy,draw=morange!90!black,fill=morange!12,text width=43mm] (pt) at (10.25,0.35)
  {\textbf{Traversal \& Stop Policy}\\[-1pt]\scriptsize depth $\cdot$ width $\cdot$ anchor\\[-1pt]\scriptsize connector $\cdot$ low-gain stop};
\node[policy,draw=mteal!90!black,fill=mteal!10] (pe) at (15.25,0.35)
  {\textbf{Evidence Policy}\\[-1pt]\scriptsize budget $\cdot$ floor $\cdot$ reservation};

\coordinate (bus) at (7.35,1.42);
\draw[mblue,line width=0.85pt] (pi.south) -- (bus);
\draw[mblue,line width=0.85pt] (bus) -| (ps.north);
\draw[mblue,line width=0.85pt] (bus) -| (pt.north);
\draw[mblue,line width=0.85pt] (bus) -| (pe.north);

\node[stage,draw=mblue,fill=mblue!4] (s) at (5.25,-2.05)
  {\textbf{Seed Selection}};
\node[stage,draw=morange!90!black,fill=morange!8,text width=43mm] (t) at (10.25,-2.05)
  {\textbf{Path Traversal}\\[-1pt]\scriptsize adaptive stopping during exploration};
\node[stage,draw=mteal!90!black,fill=mteal!6] (e) at (15.25,-2.05)
  {\textbf{Evidence Selection}};
\node[stage,draw=black!55,fill=mgray,text width=34mm] (g) at (19.75,-2.05)
  {\textbf{Fixed Answer}\\\textbf{Generation}};

\draw[control,mblue] (ps) -- (s);
\draw[control,morange!90!black] (pt) -- (t);
\draw[control,mteal!90!black] (pe) -- (e);
\draw[arr] (s) -- (t);
\draw[arr] (t) -- (e);
\draw[arr] (e) -- (g);

\node[draw=black!45,dashed,rounded corners=3pt,fit=(s)(t)(e),inner xsep=4mm,inner ysep=4mm,
  label={[label]below:Shared corpus graph, indexes, scoring, and retrieval implementation}] {};
\node[label,above=1.5mm of g] {not policy-controlled};
\end{tikzpicture}
}
\caption{\method\ maps a query to one validated policy with stage-specific controls. Seed, traversal and stopping, and evidence controls configure a shared GraphRAG pipeline, while answer generation remains fixed.}
\label{fig:overview}
\end{figure}

Given a policy, the complete computation is
\begin{align}
S_q &= \mathcal{S}(G,q;\pi_q^{\mathrm{seed}}),\\
P_q &= \mathcal{T}(G,S_q;\pi_q^{\mathrm{trav}}),\\
\widetilde E_q &= \mathcal{F}(q,P_q;\pi_q^{\mathrm{filter}}),\\
\widehat y_q &= \mathcal{M}_{\mathrm{ans}}(q,\operatorname{Ground}(\widetilde E_q)).
\end{align}

\subsection{Policy space and validation}

The analyzer starts from conservative defaults and overrides a control only when the question provides a clear signal. \Cref{tab:policyspace} reports the paper-run interface.

\begin{table}[H]
\centering
\caption{Bounded policy controls.}
\label{tab:policyspace}
\small
\begin{tabularx}{\textwidth}{@{}l p{0.31\textwidth} l X@{}}
\toprule
Control & Valid values & Default & Function \\
\midrule
\code{seed\_limit} & integer 1--15 & 7 & Deduplicated seed budget \\
\code{seed\_query\_mode} & full query, compact keywords, per-target keywords & full query & Seed-query construction \\
\code{selection\_mode} & relation-only, hybrid & relation-only & Entity/relation contribution \\
\code{max\_depth} & integer 2--5 & 3 & Hard traversal cap \\
\code{beam\_width} & integer 4--12 & 8 & Partial paths retained per depth \\
\code{anchor\_strength} & real 0--1 & 0.0 & Target focus and hub suppression \\
\code{use\_connector} & Boolean & false & Cross-seed connector exploration \\
\code{stop\_regime} & shallow, medium, deep & shallow & Low-gain sensitivity \\
\code{evidence\_count} & integer 5--18 & 8 & Final evidence cap \\
\code{min\_evidence} & integer 1--12 & 3 & Evidence preservation floor \\
\code{adaptive\_final\_evidence} & Boolean & false & Role-aware path reservation \\
\bottomrule
\end{tabularx}
\end{table}

The analyzer also returns target entities, target facets, a short retrieval-risk description, and a rationale for diagnostics. These fields do not bypass policy validation. Unsupported categorical values are rejected, numeric fields are clamped, and missing values revert to defaults. This ensures that the LLM controls a finite interface rather than directly executing arbitrary code or inventing new retrieval operators.

\subsection{Seed selection}

The selected query mode searches fixed entity and relation indexes with the complete question, a compact query, or one query per extracted target. Let $s_{\mathrm{ent}}(v,q_e)$ and $s_{\mathrm{rel}}(v,q_e)$ denote entity- and relation-index scores for node $v$. Relation-only selection uses
\begin{equation}
s_{\mathrm{seed}}(v,q)=s_{\mathrm{rel}}(v,q_e),
\end{equation}
while hybrid selection uses
\begin{equation}
s_{\mathrm{seed}}(v,q)=0.3s_{\mathrm{ent}}(v,q_e)+0.7s_{\mathrm{rel}}(v,q_e).
\end{equation}
The seed set is $S_q=\operatorname{TopK}_{K_q}\{v:s_{\mathrm{seed}}(v,q)\}$. In per-target mode, the two best candidates for each target are preserved before node-level deduplication and final truncation. This prevents a comparison from allocating all seeds to its most salient entity.

\subsection{Policy-conditioned traversal}

Let $p=(v_0,e_1,v_1,\ldots,e_h,v_h)$ be a path with $h$ edges. The structure-aware score combines relation similarity $\bar r$, target relevance $\bar t$, specificity $\bar s$, anchor consistency $\bar a$, transition alignment $\bar x$, path coherence $\bar c$, hub exposure $\bar h$, repetition $\rho_{\mathrm{rep}}$, and question-like labels $\rho_{\mathrm{ql}}$:
\begin{equation}
s_{\mathrm{path}}(p,q)=w_r\bar r+w_t\bar t+w_s\bar s+w_a\bar a+w_x\bar x+w_c\bar c-w_h\bar h-w_{\mathrm{rep}}\frac{\rho_{\mathrm{rep}}}{h}-w_{\mathrm{ql}}\rho_{\mathrm{ql}}.
\label{eq:pathscore}
\end{equation}
Default weights are $(0.45,0.15,0.15,0.10,0.15,0.05,0.20,1.0,1.0)$. When $a_q\geq0.4$, target, anchor, transition, and hub terms increase with $a_q$, changing a general relation-oriented scorer into a target- and transition-aware scorer. At depth $d$, all retained paths are expanded, connector-view candidates are added when enabled, and the best $W_q$ paths are kept. $D_q$ remains the hard cap.

Connector exploration is a second view, not a separate pipeline. Standard beam expansion grows outward from each seed; the connector view searches for paths that join regions initialized by distinct targets. Both views enter the same scoring and pruning pool.

\subsection{Cumulative-gain stopping}

Let $A^N_{d-1}$ and $A^E_{d-1}$ be nodes and edges accumulated before depth $d$, and $N_d,E_d$ the structure reached at the current depth. Novelty gain is
\begin{equation}
g_d=\frac{|N_d\setminus A^N_{d-1}|+|E_d\setminus A^E_{d-1}|}{\max(1,|A^N_{d-1}|+|A^E_{d-1}|)}.
\end{equation}
After minimum depth $d_{\min}=2$, a step is eligible for early stopping when $g_d\leq\epsilon_r$ and cumulative evidence completeness $c_d\geq0.35$. With patience one,
\begin{equation}
\operatorname{Stop}(q,d)=\mathbb{I}[d\geq D_q]\lor\mathbb{I}[d\geq2\land g_d\leq\epsilon_{r_q}\land c_d\geq0.35].
\end{equation}
The paper-run thresholds are 5.00, 0.35, and 0.24 for shallow, medium, and deep regimes. The large shallow threshold intentionally stops at the first completeness-eligible depth; it does not mean novelty is numerically small in an absolute sense.

\subsection{Evidence selection and source grounding}

After traversal, paths are deduplicated and ranked. A shared path filter operates on at most ten candidates. When adaptive reservation is disabled, selection uses the global top paths. When enabled, it first preserves up to four core paths, then reserves up to four additional positions for complementary roles: a connector path, an uncovered target, an uncovered facet, or a previously unseen first-relation label. Remaining positions are filled by path score.

The ranked paths are flattened into deduplicated evidence units. $B_q$ caps retained evidence, while $L_q$ restores high-ranked excluded units when selection falls below the preservation floor. Importantly, the evaluated \method\ configuration does not add a separate LLM call for post-traversal noise filtering. Noise is controlled through policy-conditioned traversal, path scoring, cumulative-gain stopping, role-aware reservation, and local source-window reranking.

Graph evidence is mapped back to source chunks. Candidate chunks are divided into 1,600-character windows with a stride of 800. A local BAAI/bge-large-en-v1.5 bi-encoder ranks the windows by cosine similarity to the question, and the top 15 windows are passed to GPT-4o-mini at temperature zero. This preserves graph structure during exploration while returning full textual detail for answer generation.

\begin{tcolorbox}[colback=white,colframe=mblue,title=Algorithm 1: MOSAIC query-specific graph retrieval]
\small
\textbf{Input:} query $q$, corpus graph $G$, entity and relation indexes.\quad
\textbf{Output:} answer $\hat y_q$, grounded windows $W_q$.
\begin{enumerate}
  \item $z_q\gets\textsc{AnalyzeRequirements}(q)$; $\pi_q\gets\textsc{ValidateAndInstantiate}(z_q)$.
  \item $S_q\gets\textsc{RetrieveSeeds}(q,\pi_q^{\mathrm{seed}})$; initialize $\mathcal{B}_0\gets S_q$ and $P_q\gets\varnothing$.
  \item For $d=1,\ldots,\pi_q.D_{\max}$: expand $\mathcal{B}_{d-1}$; rank and prune with $\pi_q^{\mathrm{trav}}$; merge retained paths into $P_q$; stop when $d\geq2$ and the policy-conditioned stopping test succeeds.
  \item $E_q\gets\textsc{SelectEvidence}(P_q,q,\pi_q^{\mathrm{filter}})$. If $|E_q|<\pi_q.L$, restore the highest-ranked excluded evidence up to the floor.
  \item $W_q\gets\textsc{GroundToSource}(E_q,q)$; $\hat y_q\gets\textsc{GenerateAnswer}(q,W_q)$; return $(\hat y_q,W_q)$.
\end{enumerate}
\end{tcolorbox}

\section{Experimental Setup}

\paragraph{Primary benchmarks.} GraphRAG-Bench Medical contains 2,062 questions over 2,406 medical-guideline documents; Novel contains 2,010 questions over 461 literary documents. Both contain Fact Retrieval (FR), Complex Reasoning (CR), Contextual Summarization (CS), and Creative Generation (CG). We use every question and the original corpora.

\paragraph{Metrics.} Answer Correctness (ACC) is the primary end-to-end metric because it is defined for all four categories. Overall ACC is query-count weighted, not the unweighted mean of category scores. Retrieval is evaluated with Evidence Recall (ER) and Context Relevancy (CR). Task-specific ROUGE-L, coverage, and faithfulness are reported as secondary metrics.

\paragraph{Implementation.} Each corpus graph is constructed once with a LightRAG-style entity--relation extraction pipeline. OpenAI text-embedding-3-large supplies semantic representations. GPT-4o-mini at temperature 0 is used for the analyzer and answer generation. The graph, indexes, retrieval code, and generator are shared by \method\ and fixed-policy controls. \method\ requires no additional training or fine-tuning.

\paragraph{Baselines.} We compare with reported GraphRAG-Bench systems and with AutoPrunedRetriever and G-Reasoner. Because published baselines were not rerun in our implementation, claims involving them are comparisons to externally reported values. To isolate policy adaptation, we additionally implement Fixed Narrow, Medium, Wide, and a Budget-Matched Fixed policy on the identical graph and generator.

\section{Results}

\subsection{Answer correctness}

\begin{table}[t]
\centering
\caption{Answer Correctness (\%) on GraphRAG-Bench. Overall is query-count weighted. Best values are bold.}
\label{tab:mainacc}
\small
\resizebox{\textwidth}{!}{%
\begin{tabular}{lccccc|ccccc}
\toprule
& \multicolumn{5}{c|}{Medical} & \multicolumn{5}{c}{Novel}\\
Method & Overall & FR & CR & CS & CG & Overall & FR & CR & CS & CG\\
\midrule
RAG w/ rerank & 63.04&64.73&58.64&65.75&60.61&52.97&60.92&42.93&51.30&38.26\\
HippoRAG2 &64.91&66.28&61.98&63.08&68.05&58.41&60.14&53.38&64.10&48.28\\
AutoPruned (LLM)&65.35&61.25&71.59&70.14&65.02&58.34&45.99&62.80&\textbf{83.10}&\textbf{62.97}\\
G-Reasoner&71.84&68.84&75.17&77.23&\textbf{72.04}&59.90&60.07&53.92&71.28&50.48\\
\midrule
\textbf{\method}&\textbf{76.97}&\textbf{76.05}&\textbf{76.92}&\textbf{85.28}&68.78&\textbf{64.33}&\textbf{66.06}&57.21&73.12&56.54\\
\bottomrule
\end{tabular}}
\end{table}

\method\ obtains 76.97 on Medical and 64.33 on Novel, improving over the strongest previously reported overall results by 5.13 and 4.43 points. On Medical it leads FR, CR, and CS; it does not lead CG. On Novel, it leads overall and FR but trails specialized systems on CR, CS, and CG. We therefore interpret the result as consistent overall accuracy across heterogeneous evidence requirements, not uniform dominance on every task category.

\subsection{Retrieval quality}

\begin{table}[t]
\centering
\caption{Retrieval performance (\%). ER: Evidence Recall; CR: Context Relevancy.}
\label{tab:retrieval}
\small
\begin{tabular}{lcccc}
\toprule
&\multicolumn{2}{c}{Medical}&\multicolumn{2}{c}{Novel}\\
Method&ER&CR&ER&CR\\
\midrule
RAPTOR&84.2&62.6&66.1&58.0\\
LightRAG&82.6&42.2&79.6&35.5\\
HippoRAG2&73.6&85.3&66.2&\textbf{82.8}\\
G-Reasoner&93.8&--&87.7&--\\
\midrule
\textbf{\method}&\textbf{95.1}&\textbf{86.1}&\textbf{90.2}&77.3\\
\bottomrule
\end{tabular}
\end{table}

\method\ achieves the highest Evidence Recall on both domains, exceeding G-Reasoner by 1.3 points on Medical and 2.5 points on Novel. It also achieves the highest reported Medical Context Relevancy. Novel reveals a useful trade-off: HippoRAG2 has higher Context Relevancy, but substantially lower Evidence Recall.

\subsection{Why one fixed policy is insufficient}

\begin{table}[t]
\centering
\caption{Controlled fixed-policy comparison on Medical. All systems share graph and generator.}
\label{tab:fixed}
\small
\begin{tabular}{lrrrrrrrrr}
\toprule
Policy&Seeds&Depth&Width&Evidence&FR&CR&CS&CG&Overall\\
\midrule
Narrow&3&2&4&1--3&.620&.674&.733&.686&.6545\\
Medium&5&4&8&5--8&.636&.683&.758&\textbf{.703}&.6701\\
Wide&10&5&16&12--15&.630&.672&.765&.683&.6635\\
\midrule
\method&\multicolumn{4}{c}{adaptive per query}&\textbf{.761}&\textbf{.769}&\textbf{.853}&.688&\textbf{.7697}\\
\bottomrule
\end{tabular}
\end{table}

The relationship between scope and accuracy is non-monotonic. Medium improves on Narrow, but Wide declines despite more seeds, deeper search, a larger beam, and more evidence. Fixed Wide helps CS but hurts FR, CR, and CG relative to Medium. Benchmark category also does not uniquely determine the correct policy: some fact questions require mediated evidence, while some summaries are supported by a compact neighborhood. \method\ exceeds the strongest canonical fixed policy by 9.96 points, showing that the key benefit is query-level allocation rather than choosing a single larger operating point.

\subsection{Policy diversity and requirement alignment}

Across 2,062 Medical queries, the analyzer produces 19 unique policy combinations. Assigned depth is distributed across 2, 3, and 4 hops for 26\%, 43\%, and 31\% of queries. Seed count ranges from 5 to 12; evidence budgets range from 8 to 16. Coupled connector/two-view exploration is activated for 32\%, adaptive evidence selection for 35\%, hybrid identity--relation scoring for 19\%, and per-target seeding for 19\%. This rules out collapse to a single dominant configuration.

\begin{figure}[t]
\centering
\begin{subfigure}{0.49\textwidth}
\includegraphics[width=\linewidth]{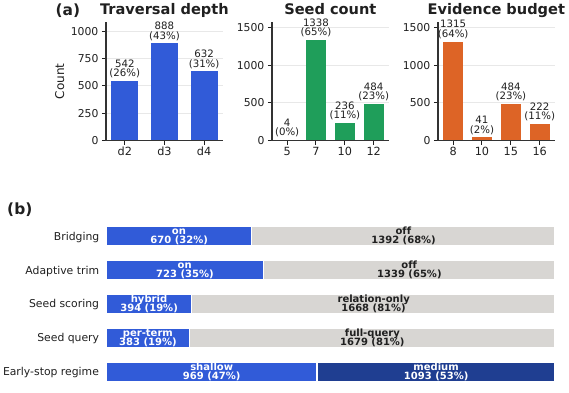}
\caption{Policy diversity.}
\end{subfigure}\hfill
\begin{subfigure}{0.49\textwidth}
\includegraphics[width=\linewidth]{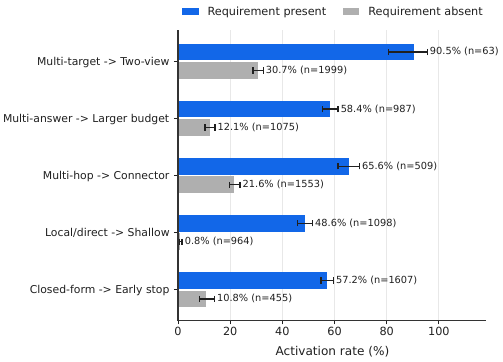}
\caption{Requirement--behavior alignment.}
\end{subfigure}
\caption{Analyzer outputs on 2,062 Medical questions. Error bars in (b) are 95\% Wilson intervals.}
\label{fig:policyanalysis}
\end{figure}

Independent post-hoc proxies further show that variation is purposeful. Two-view traversal is activated for 90.5\% of multi-target queries versus 30.7\% of others; a large evidence budget for 58.4\% of multi-answer queries versus 12.1\%; connector traversal for 65.6\% of multi-hop queries versus 21.6\%; shallow traversal for 48.6\% of local/direct queries versus 0.8\%; and early-stop behavior for 57.2\% of closed-form queries versus 10.8\%. The proxies use deterministic surface, gold-answer-structure, and benchmark-label rules only for analysis; they are never provided to the analyzer.

\subsection{Diagnostic counterfactuals}

\begin{table}[t]
\centering
\caption{Queries improved by the complete policy relative to disabling the corresponding execution control. Counts are diagnostic and non-additive.}
\label{tab:counterfactual}
\small
\begin{tabular}{lr@{\qquad}lr}
\toprule
Consideration&Improved&Consideration&Improved\\
\midrule
Mediatedness&48&Output cardinality&190\\
Seed coverage&55&Anchoring need&46\\
Identity--relation dependence&41&Completeness&47\\
\bottomrule
\end{tabular}
\end{table}

These counterfactuals measure the operational effect of a control while retaining the full analyzer. They should not be summed: the same query can improve through several interacting controls. The strongest single count belongs to output cardinality, consistent with the tendency of high-scoring paths to crowd out sibling evidence needed for list-like answers.

\subsection{Latency and graph-search effort}

\begin{table}[t]
\centering
\caption{Accuracy and average per-query latency on Medical. ACC uses all 2,062 queries; latency uses the same 197-query subset.}
\label{tab:latency}
\small
\begin{tabular}{lrrrrr}
\toprule
Method&ACC&Retrieval&Analyzer&Generation&End-to-end\\
\midrule
Fixed Narrow&65.45&3.27&--&1.76&5.03\\
Fixed Medium&67.01&3.50&--&1.79&5.29\\
Fixed Wide&66.35&4.81&--&1.91&6.72\\
Budget-matched Fixed&72.03&3.63&--&2.26&5.89\\
\method&76.97&4.67&2.82&2.19&9.68\\
\bottomrule
\end{tabular}
\end{table}

\begin{table}[t]
\centering
\caption{Graph-search effort relative to Fixed Wide.}
\label{tab:effort}
\small
\begin{tabular}{lrrr}
\toprule
Metric&Fixed Wide&\method&Reduction\\
\midrule
Evaluated paths&800&145&81.9\%\\
Reached depth&5.00&2.59&48.2\%\\
Beam width&16.0&8.0&50.0\%\\
Selected paths&16.0&9.4&41.3\%\\
Retained evidence items&12.35&6.52&47.2\%\\
\bottomrule
\end{tabular}
\end{table}

\method\ is not faster end-to-end in the current implementation. Its separate policy-construction call adds 2.82 seconds on average, producing 9.68 seconds total versus 5.29 for Fixed Medium. The relevant efficiency result is different: additional LLM computation is used to allocate much less graph search more effectively. The analyzer and seed-keyword extraction are currently separate and unbatched; caching, prompt compression, call fusion, or distillation could reduce control latency without changing the retrieval interface.

\section{Transfer to Multi-Hop QA}

We evaluate the same policy space on 1,000-question subsets of HotpotQA, MuSiQue, and 2WikiMultiHopQA. No benchmark training split or task-specific retriever fine-tuning is used. One or two short answer-format instructions are added for each benchmark.

\begin{table}[t]
\centering
\caption{Transfer results (\%). G-Reasoner values are externally reported and use target-benchmark training; \method\ is not trained on these benchmarks.}
\label{tab:transfer}
\small
\begin{tabular}{lrrrrrr}
\toprule
&\multicolumn{2}{c}{\method}&\multicolumn{2}{c}{G-Reasoner}&\multicolumn{2}{c}{Difference}\\
Dataset&EM&F1&EM&F1&EM&F1\\
\midrule
HotpotQA&61.1&75.3&61.4&76.0&-0.3&-0.7\\
MuSiQue&40.3&51.2&38.5&52.5&+1.8&-1.3\\
2WikiMultiHopQA&62.3&70.0&74.9&82.1&-12.6&-12.1\\
\bottomrule
\end{tabular}
\end{table}

\method\ is close to the trained reference on HotpotQA and exceeds it by 1.8 EM on MuSiQue, while a substantial gap remains on 2WikiMultiHopQA. A separate LLM-based semantic correctness evaluation yields 77.9, 51.3, and 75.1, suggesting that lexical EM/F1 penalize some semantically correct but differently formatted answers. Because the systems were not rerun under a common pipeline and G-Reasoner uses benchmark-specific supervision, these results support transfer but do not establish a state-of-the-art claim.

\section{Qualitative Analysis}

\begin{table}[t]
\centering
\caption{Representative controlled cases from Medical. Percentages denote ACC.}
\label{tab:cases}
\small
\begin{tabularx}{\textwidth}{@{}p{0.18\textwidth}p{0.30\textwidth}XX@{}}
\toprule
Case&Question&Outcome&Retrieval explanation\\
\midrule
Bridge recovery&How are multiple tumors in the kidneys classified in relation to metastasis?&\method\ 100\%; Narrow 22\%; Wide 23\%.&Depth 5, 12 seeds, and connector exploration recover the path through \emph{Primary Tumor}; anchoring prevents staging evidence from dominating.\\
Drift prevention&What initial diagnostic step is critical for an anterior mediastinal mass in a person assigned male at birth?&\method\ 100\%; Wide 23\%.&Wide drifts through a demographic phrase into breast-cancer evidence. Target anchoring retains the MGZL diagnostic context.\\
Multi-answer completeness&What are the most common subtypes of glioma?&\method\ 100\%; fixed policies 80\%.&Role-aware reservation preserves astrocytoma, oligodendroglioma, and glioblastoma instead of letting grade-related paths crowd out sibling subtype relations.\\
Analyzer over-expansion&What is the role of sentinel lymph node biopsy in CSCC?&\method\ 24\%; Narrow 99\%.&The analyzer incorrectly predicts mediation; depth-5 connector exploration introduces generic surgery and care-team nodes. This is a policy-calibration failure, not graph unreachability.\\
\bottomrule
\end{tabularx}
\end{table}

The successful cases show why controls must be combined. More depth alone does not recover the kidney-tumor answer because the larger context can still be dominated by a nearby but incorrect staging interpretation. The failure case is equally important: an expressive controller can overreact. Guardrails constrain the output range, but they cannot guarantee that the evidence requirement itself is inferred correctly.

\section{Discussion}

\subsection{What the results establish}

The controlled experiments establish that query-specific policy adaptation improves over several globally fixed operating points on a shared graph and generator. Policy logs show that the controller neither collapses to one configuration nor varies arbitrarily: activations align with independent requirement proxies. The graph-search measurements show that gains are not produced by uniformly retrieving more evidence.

\subsection{What the results do not establish}

The experiments do not isolate a causal contribution for every prompt phrase or every low-level scoring term. Controls interact, and the counterfactual counts are not additive. Comparisons to published GraphRAG systems are not component-matched reproductions. Transfer results compare against external G-Reasoner numbers under asymmetric training conditions. Finally, the current latency reflects an unoptimized separate analyzer call.

\subsection{Extensibility and deployment}

The bounded policy schema offers a practical separation of concerns. Retrieval engineers define safe controls and defaults; the analyzer maps natural-language evidence requirements to those controls; monitoring uses policy logs and retrieval traces. New failure patterns can be addressed without introducing a new end-to-end pipeline for each query type. In production, policies can be cached for recurring query structures, distilled into a smaller controller, or constrained further by domain rules. Because the answer model consumes grounded source windows and remains outside the controller's free-form action space, the framework also supports clearer auditing of how a query changed retrieval.

\section{Limitations}

First, the analyzer is an LLM and can misclassify evidence structure, as the SLNB case demonstrates. Second, the current signals and valid ranges were developed from observed failures in the evaluated setting; transfer to different graph schemas may require additional calibration. Third, graph construction quality bounds retrieval: missing or incorrectly merged entities and relations cannot be repaired solely by a better policy. Fourth, answer quality still depends on source grounding and generation, particularly for open-ended CG questions where \method\ does not uniformly improve. Fifth, latency and monetary cost are higher than fixed retrieval in the current implementation. Finally, some comparisons rely on externally reported baselines, and a fully component-matched reproduction across all methods remains future work.

\section{Conclusion}

\method\ treats GraphRAG retrieval as a per-query control problem. An analyzer converts evidence requirements into a bounded policy spanning seed selection, traversal, stopping, and evidence retention, while the corpus graph and answer generator remain shared. The approach improves overall answer correctness and evidence recall on GraphRAG-Bench and uses substantially less graph search than a uniformly wide policy. Its six current considerations are composable implementation signals, not a fixed taxonomy. The broader result is that effective GraphRAG should adapt not merely relevance scores, but the operating policy of graph exploration itself.

\section*{Artifact and Reproducibility Statement}

The evaluated implementation contains proprietary components. For result verification, the authors intend to provide benchmark configurations, dependencies, graph construction and retrieval code required for reproduction, inference scripts, generated outputs, and official evaluation commands through a controlled-access repository, subject to organizational approval and applicable benchmark licenses.

\end{document}